\documentclass[sigconf]{acmart}

\usepackage{microtype}
\usepackage{graphicx}
\usepackage{subfigure}
\usepackage{booktabs}
\usepackage{multirow}
\usepackage{subcaption}
\usepackage{enumitem}
\usepackage{colortbl}
\usepackage{tablefootnote}

\AtBeginDocument{%
  }

\acmYear{2025}
\usepackage{pifont}
\newcommand{\cmark}{\ding{51}} 
\newcommand{\xmark}{\ding{55}}
\begin{document}

\title{Empowering Time Series Analysis with Synthetic Data: A Survey and Outlook in the Era of Foundation Models}


\author{Xu Liu}
\thanks{Junnan Li and Chenghao Liu are corresponding authors.}
\affiliation{%
  \institution{Salesforce AI Research \& National University of Singapore}
  \country{}}
\email{liuxu726@gmail.com}

\author{Taha Aksu}
\affiliation{%
  \institution{Salesforce AI Research}
  \country{}}
\email{iaksu@salesforce.com}

\author{Juncheng Liu}
\affiliation{%
  \institution{Salesforce AI Research}
  \country{}}
\email{juncheng.liu@salesforce.com}

\author{Qingsong Wen}
\affiliation{%
  \institution{Squirrel Ai Learning}
  \country{}}
\email{qingsongedu@gmail.com}

\author{Yuxuan Liang}
\affiliation{%
  \institution{Hong Kong University of Science and Technology (Guangzhou)}
  \country{}}
\email{yuxliang@outlook.com}

\author{Caiming Xiong}
\affiliation{%
  \institution{Salesforce AI Research}
  \country{}}
\email{cxiong@salesforce.com}

\author{Silvio Savarese}
\affiliation{%
  \institution{Salesforce AI Research}
  \country{}}
\email{ssavarese@salesforce.com}

\author{Doyen Sahoo}
\affiliation{%
  \institution{Salesforce AI Research}
  \country{}}
\email{dsahoo@salesforce.com}

\author{Junnan Li}
\author{Chenghao Liu}
\affiliation{%
  \institution{Salesforce AI Research}
  \country{}}
\email{{junnan.li,chenghao.liu}@salesforce.com}

\renewcommand{\shortauthors}{Xu Liu et al.}

\begin{abstract}
Time series analysis is crucial for understanding dynamics of complex systems. Recent advances in foundation models have led to task-agnostic Time Series Foundation Models (TSFMs) and Large Language Model-based Time Series Models (TSLLMs), enabling generalized learning and integrating contextual information. However, their success depends on large, diverse, and high-quality datasets, which are challenging to build due to regulatory, diversity, quality, and quantity constraints. Synthetic data emerge as a viable solution, addressing these challenges by offering scalable, unbiased, and high-quality alternatives. This survey provides a comprehensive review of synthetic data for TSFMs and TSLLMs, analyzing data generation strategies, their role in model pretraining, fine-tuning, and evaluation, and identifying future research directions.
\end{abstract}




\maketitle

\section{Introduction}
Time series analysis is essential for revealing the underlying dynamics of complex systems and processes \cite{hamilton2020tsanalysis,lim2021tsforecastsurvey}. Essential tasks within this field, such as forecasting~\cite{liu2022contrastive,zhou2023ofa,cao2023tempo}, classification~\cite{feofanov2025mantis,liu2024diffshape}, and anomaly detection~\cite{chen2024mace,chen2024lara}, are vital for informed decision-making, identifying patterns, and detecting irregularities. With its far-reaching practical applications, time series analysis has long been a focal point of research and innovation, continuously drawing significant interest from both academia and industry.

In recent years, breakthroughs in foundation models (FMs) for language and vision have driven a transformative shift in time series analysis \cite{liang2024foundation}. The traditional paradigm of building specialized models for specific tasks/datasets \cite{wen2022transformers,zhou2022fedformer,woo2022etsformer,nie2022time} has transited to task-agnostic time series foundation models (TSFMs) \cite{woo2024moirai,liu2024moiraimoe,liu2024timerxl,rasul2023lagllama,zukowska2024icm,garza2023timegpt,zhang2024elastst}. The power of TSFMs lies in their ability to harness large-scale data, enabling the creation of generalized representations that can be directly applied in a zero-shot setting or refined through finetuning across diverse downstream tasks ~\cite{zhang2024large}. Meanwhile, the remarkable advancements in Large Language Models (LLMs) have enabled the integration of rich contextual information beyond raw numerical data in time series analysis. This progress has given rise to novel models called time series LLMs or TSLLMs \cite{xie2024chatts,jia2024gpt4mts,liu2024unitime,jin2023timellm}, and novel tasks such as time series reasoning~\cite{chow2024tsreason,ye2024tsreasoner}, captioning~\cite{emami2024syscaps,zhang2023insightminer}, and question answering~\cite{wang2025chattime,cai2024timeseriesexam}, making interactive and interpretable time series analytics a tangible reality.

However, the success of both TSFMs and TSLLMs depends on access to large, diverse, and high-quality datasets for pretraining and evaluation, and acquiring such datasets remain a significant challenge \cite{liu2024best}. First, while abundant third-party time series data sources exist, their use must comply with legal and regulatory frameworks, imposing significant restrictions on the commercial deployment of pretrained models ~\cite{liu2024time}. Second, collected datasets often suffer from limited diversity and inherent biases—stemming not only from the natural imbalance between low- and high-frequency time series but also because of variations in how researchers design collection pipelines \cite{ismail2019deep}. Data quality is also not guaranteed, with data sometimes containing long periods of missing values, excessive noise and outliers, complicating preprocessing and integration \cite{goyal2024systematic}. Third, a unique issue with TSLLMs is that paired time series-text data remain scarce in real-world settings, posing a significant bottleneck for emerging tasks like time series reasoning \cite{xie2024chatts}.

To mitigate the issues mentioned above, synthetic data have emerged as a viable alternative \citep{nikitin2025tsgm}. First, synthetic time series can be easily generated at scale without license constraints, ensuring a plentiful supply of pretraining and evaluation data, and the large-scale deployment of pretrained models. This is also beneficial in privacy-sensitive domains like healthcare and finance, where access to real data is often restricted. Second, synthetic data allow for better control over data quality and diversity, which can reduce bias from human and enhance model generalization. For example, synthetic data play a vital role in supplementing real data during the pretraining of TSFMs like Chronos \cite{ansari2024chronos} and TimesFM \cite{das2024timesfm}, enriching learned patterns and significantly boosting downstream performance. Third, the scarcity of time series-text data pairs \cite{chow2024tsreason,liu2024timemmd} has been effectively alleviated due to the use of generative tools like LLMs to annotate time series with contextual descriptions. This process facilitates the alignment of both modalities and accelerates the development of TSLLMs for time series understanding and reasoning.

\begin{table}[!t]
    \centering
    \caption{Comparison between our survey and related surveys.}
    \vspace{-1em}
    \label{tab:survey-compare}
    \resizebox{0.9\linewidth}{!}{
    \begin{tabular}{ccccccc}
        \toprule
        Survey & Year &TSLLMs & TSFMs & Syn Data Generation & Syn Data Usage \\
        \midrule
        \citet{jin2023large} &2023  & \cmark &\cmark & \xmark & \xmark \\
        \citet{jiang2024empowering} & 2024 & \cmark &\xmark & \xmark & \xmark \\
        \citet{zhang2024large} & 2024 & \cmark &\xmark & \xmark & \xmark \\
        \citet{liang2024foundation} &2024  & \cmark &\cmark & \xmark & \xmark \\
        \citet{long2024llms} &2024 & \xmark & \xmark & \cmark & \xmark \\
        \citet{tan2024large} &2024 & \xmark & \xmark & \cmark & \cmark \\
        \midrule
        \midrule
        (Ours) &2025 & \textbf{\cmark} & \textbf{\cmark} & \textbf{\cmark} & \textbf{\cmark} \\
        \bottomrule
    \end{tabular}
    }
    \vspace{-1em}
\end{table}

While synthetic data have proven invaluable, a comprehensive analysis of their generation and application in TSFMs and TSLLMs remains unexplored. As summarized in Table \ref{tab:survey-compare}, prior studies primarily focus on methodologies or downstream applications of TSFMs and TSLLMs \cite{jiang2024empowering,zhang2024large,liang2024foundation,jin2023large}, without examining them from a data-centric perspective. Meanwhile, surveys such as \citet{long2024llms} and \citet{tan2024large} delve into synthetic data generation but remain confined to the language domain, leaving time series largely unaddressed. To bridge this gap, this paper aims to provide a comprehensive and up-to-date survey regarding synthetic data for TSFMs and TSLLMs. Specifically, we propose to structure this survey around TSFMs and TSLLMs, as each type of model requires distinct data generation strategies, utilizes different foundation models, and serves unique downstream applications. Within each model type, we analyze existing methodologies through the lens of the model development lifecycle, tracking the progression from synthetic data generation to their application in key stages, including pretraining, finetuning, and evaluation. Each section concludes with a discussion of the limitations, and finally we highlight potential future research directions for both TSFMs and TSLLMs.

\section{Preliminaries}
\subsection{Time Series Analytical Tasks}
In this part, we provide a brief introduction to the time series analytical tasks covered in this survey. For TSFMs, we focus on the two most widely adopted downstream tasks: given an input time series, \textit{forecasting} predicts a sequence of future numerical values, while \textit{classification} assigns the series to a predefined option based on its patterns. Additionally, we identify five broad categories of downstream tasks for TSLLMs. \textit{Question Answering (QA)} requires generating a textual answer given a time series and a natural language question, either focusing on statistical properties or structural patterns. \textit{Multiple Choice Question Answering (MCQA)} follows a similar format but restricts responses to predefined options, often testing a model’s ability to recognize patterns or infer relationships. \textit{Reasoning} involves higher-order cognitive tasks such as causal inference (identifying underlying factors influencing a time series), comparative reasoning (contrasting attributes across different series), and deductive reasoning (applying logical rules to interpret patterns). Across the literature, reasoning tasks are structured as QA, MCQA, or open-ended text generation given a time series input. \textit{Captioning} entails generating a textual summary of a time series. \textit{Forecast Explanation (FE)} focuses on explaining forecasting model predictions by providing insights into expected trends, past influences, or sources of uncertainty.

\subsection{Generation of Synthetic Data} 
\label{sec:preliminary_generation}
Synthetic data generation can supply vast amounts of artificial data for model training and evaluation, playing an increasingly vital role in the foundation model era. The generation of synthetic time series can be broadly classified into three categories ~\cite{nikitin2025tsgm,ang2023tsgbench}. \textit{Statistical} and \textit{simulator-based} approaches generate time series through predefined rules or simulation environments, respectively, allowing for controlled and interpretable data creation. In contrast, \textit{data-driven} methods leverage historical data and generative models, including GANs~\cite{yoon2019timegan,esteban2017rgan}, VAEs~\cite{desai2021timevae,lee2023timevqvae}, diffusion models~\cite{tashiro2021csdi,narasimhan2024timeweaver} or LLMs \cite{aksu2024xforecast,ye2024tsreasoner}, to learn complex temporal patterns and produce more realistic time series. Earlier methods to text generation primarily used statistical language models to model the conditional probabilities of words based on an n-gram context. However, data-driven methods, particularly LLMs, have since become the dominant approach \cite{li2024pre}.

\section{Synthetic Data for Time Series FMs}
\label{sec:tsfm}
Instead of developing separate models for each task or dataset, time series analysis is shifting toward TSFMs \cite{liu2024timer,zhang2024up2me,gao2024units,kamarthi2023lptm,chen2024visionts,feng2024gtt,ekambaram2024ttms}, which learn unified representations for generalization across tasks and domains. Synthetic data play a crucial role in TSFM development. During pretraining, it supplements patterns that real data might miss, while during evaluation, it facilitates the investigation of TSFMs' inner workings. This section reviews existing work on leveraging synthetic data for TSFMs, covering their generation and usage in model pretraining and evaluation. We then conclude by discussing the limitations of current synthetic data research. Table \ref{tab:tsfms} presents a summary of the covered works.

\subsection{Synthetic Data Generation}
Time series data is fundamentally composed of trend, seasonality, and noise, which together define its structure and dynamics. The trend represents the long-term direction of the series, showing an overall increase, decrease, or stationarity over time. Seasonality captures recurring patterns or cycles at fixed intervals, such as daily, weekly, or yearly fluctuations. Noise refers to random variations that cannot be attributed to systematic patterns, often modeled as white noise or stochastic processes. Existing synthetic time series generation methods for TSFMs mainly differ in how they define and integrate the trend, seasonality, and noise components. Notably, we identify four representative approaches that are widely adopted in follow-up studies, as shown in Table \ref{tab:ts-gen}. To provide a clearer illustration, we formally define a univariate time series (UTS) as $\mathbf{x} = \{x_1, x_2, \dots, x_T\} \in \mathbb{R}^T$, where $T$ is the sequence length. A multivariate time series (MTS) tracks multiple interrelated variables simultaneously, forming a dataset of $\mathbf{X} = \{\mathbf{x}_1, \mathbf{x}_2, \dots, \mathbf{x}_T\} \in \mathbb{R}^{N \times T}$, where $N$ denotes the number of variables.

\subsubsection{Generating Synthetic Time Series in ForecastPFN}
ForecastPFN \cite{dooley2023forecastpfn} models time series as the product of a trend and seasonal component, with an additional noise factor. The trend component $\tau(t)$, consists of both linear and exponential terms, defined by the coefficients $m_{\text{lin}}$, $c_{\text{lin}}$, $m_{\text{exp}}$, and $c_{\text{exp}}$. The seasonal component $\phi(t)$, includes periodic patterns at the week, month, and year levels. Lastly, the noise $z_t$ is characterized by the parameters $m_{\text{noise}}$ and $k$, sampled from a Weibull distribution and designed such that its expected value is 1, ensuring that, on average, it does not influence the trend or seasonality of the time series. Formally, the synthetic univariate time series $y_t$ is derived from the below equations:
\begin{align*}
    y_t & = \tau(t) \cdot \phi(t) \cdot z_t, \text{ where} \\
    z_t &= 1 + m_{\text{noise}} (z - \bar{z}), z \sim \text{Weibull}(1, k), \bar{z} = (\ln 2)^{1/k} \\
    \tau(t) &= (1 + m_{\text{lin}} \cdot t + c_{\text{lin}}) (m_{\text{exp}} \cdot c_{\text{exp}}^t) \\
    \phi(t) &= \phi_{\text{week}}(t) \cdot \phi_{\text{month}}(t) \cdot \phi_{\text{year}}(t), \text{ where} \\
    \phi_{\nu}(t) &= 1 + m_{\nu} \sum_{f=1}^{\lfloor p_{\nu}/2 \rfloor} \left[ c_{f,\nu} \sin \left( 2\pi f \frac{t}{p_{\nu}} \right) + d_{f,\nu} \cos \left( 2\pi f \frac{t}{p_{\nu}} \right) \right], \\
    \text{where} & \ \nu \in \{\text{week, month, year}\}.
\end{align*}
For all pairs of $\nu$ and $f$, the coefficients $c_{f,\nu}$ and $d_{f,\nu}$ are drawn from $\mathcal{N}(0, 1/f)$, ensuring they are inversely proportional to the harmonics of the series. These coefficients are then uniformly rescaled so that the sum of their squares equals 1. For details on the hyperparameters of $m_\nu$ and $p_\nu$, please refer to the original paper.

Subsequent works, such as Mamba4Cast~\cite{bhethanabhotla2024mamba4cast}, directly adopt the ForecastPFN algorithm, while LaT-PFN~\cite{verdenius2024latpfn} adapts and modifies ForecastPFN's time series synthetic prior. They primarily adjust the definitions of the seasonality and noise components by introducing new hyperparameters, aiming to promote coherence without oversimplifying the series and improve the model’s ability to generalize in a zero-shot setting. Another notable study, \citet{kuvshinova2024synornot}, takes a different approach. They use priors that are independent of time periodicity (such as weekly). Instead, seasonality is defined by sampling Fourier coefficients, and the trend is incorporated through a combination of various analytical functions, which are then integrated with the overall seasonal pattern.

\subsubsection{Generating Synthetic Time Series in TimesFM}
TimesFM~\cite{das2024timesfm} generates synthetic data to capture common time series patterns using statistical models. The framework is built upon four fundamental time series patterns: (1) Piecewise linear trends, where the number of segments is randomly selected between 2 and 8. (2) An autoregressive moving average (ARMA) model, parameterized by $p$ and $q$ with values ranging from 1 to 8, where the coefficients are sampled from either a multivariate Gaussian or a uniform distribution. To introduce seasonality, TimesFM incorporates (3) sine waves and (4) cosine waves with random periods between 4 and half of the maximum context length, along with random time delays. The model then randomly activates or deactivates these patterns, generates univariate time series of length 2048, and combines them using random weights sampled from a uniform distribution to construct the final synthetic dataset. Toto~\cite{cohen2024toto} adopts a synthetic data generation process similar to TimesFM to enrich its pretraining datasets and reinforce the model’s understanding of fundamental time series patterns.

\begin{table}[t]
    \LARGE
    \centering
    \caption{Synthetic time series generation comparison.}
    \vspace{-0.5em}
    \label{tab:ts-gen}%
    \resizebox{\linewidth}{!}{
    \begin{tabular}{lcccc}
        \toprule
        Representatives & Trend & Seasonality & Noise & Integration \\
        \midrule
        ForecastPFN \cite{dooley2023forecastpfn} & Linear-Exponential & Sine/Cos Waves & Weibull Distribution & Multiplication \\
        TimesFM \cite{das2024timesfm} & Piecewise Linear & Sine/Cos Waves & ARMA &  Sampling \\
        Chronos \cite{ansari2024chronos} & Linear Kernels & Periodic Kernels & RBF Kernels & GPs \\
        Moment \cite{goswami2024moment} & -- & Sine Waves & -- & -- \\
        \bottomrule
    \end{tabular}%
    }
\end{table}%

\begin{table*}[t]
    \scriptsize
    \centering
    \caption{Summary of works utilizing synthetic data for TSFMs. N.A. indicates undisclosed information. Obs: Observation. PT: Pretraining. FT: Finetuning. Eval: Evaluation. Pure: Pure synthetic data. Mixed: Synthetic data mixed with real data.}
    \vspace{-0.5em}
    \label{tab:tsfms}%
    \resizebox{\textwidth}{!}{
    \begin{tabular}{lccccccc}
        \toprule
        Method & \# TS & TS Generation & Applied Stage (Purity) & Applied Model & Downstream Task & Code & Year \\
        \midrule
        TL-TSC~\cite{rotem2022tltsc} & 15M Series & UTS (Statistical) & PT (Pure) & CNN & Classification & Yes & 2022 \\
        ForecastPFN~\cite{dooley2023forecastpfn} & 60M Obs & UTS (Statistical) & PT (Pure) & Encoder-Only Transformer & Forecasting & Yes & 2023 \\
        TimePFN~\cite{taga2024timepfn} & 2.5B Obs & MTS (Statistical) & PT (Pure) & Encoder-Only Transformer & Forecasting & No & 2024 \\
        Mamba4Cast~\cite{bhethanabhotla2024mamba4cast} & 7.3B Obs & UTS (Statistical) & PT (Pure) & Mamba-2 & Forecasting & Yes & 2024 \\
        ViTime~\cite{yang2024vitime} & 24.6M Obs & UTS (Statistical) & PT (Pure) & Vision Transformer & Forecasting & Yes & 2024 \\
        \citet{kuvshinova2024synornot} & 250M Obs & UTS (Statistical) & PT (Pure) & Encoder-only Transformer & Forecasting & Yes & 2024 \\
        LaT-PFN~\cite{verdenius2024latpfn} & 194.6M Obs & UTS (Statistical) & PT (Pure) & MobileNet-1D & Forecasting/Classification & Yes & 2024 \\
        TabPFN-TS~\cite{hoo2024tabpfnts} & 130M Series & Tabular (Statistical) & PT (Pure) & Encoder-Only Transformer & Forecasting/Classification & Yes & 2024 \\
        InfoBoost~\cite{fu2024infoboost} & 200K Series & UTS (Statistical) & PT (Pure) & BiLSTM/DLinear/PatchTST & Representation Analysis & No & 2024 \\
        
        Chronos~\cite{ansari2024chronos} & 512M Obs & UTS (Statistical) & PT (Mixed) & T5 & Forecasting & Yes & 2023 \\
        TimesFM~\cite{das2024timesfm} & 6.1B Obs & UTS (Statistical) & PT (Mixed) & Decoder-only Transformer & Forecasting & Yes & 2023 \\
        Toto~\cite{cohen2024toto} & N.A. & UTS (Statistical) & PT (Mixed) & Decoder-only Transformer & Forecasting & No & 2024 \\
        Time-MoE~\cite{shi2024timemoe} & 1B Obs & UTS (Statistical) & PT (Mixed) & Decoder-only Transformer & Forecasting & Yes & 2024 \\
        Sundial~\cite{liu2025sundial} & 512M Obs & UTS (Statistical) & PT (Mixed) & Decoder-only Transformer & Forecasting & No & 2025 \\
        TimeHF~\cite{qi2025timehf} & 98B Obs & UTS (Statistical) & PT (Mixed) & Encoder-only Transformer & Forecasting & No & 2025 \\
        
        Moment~\cite{goswami2024moment} & N.A. & UTS (Statistical) & Eval (Pure) & Moment & Representation Analysis & Yes & 2024 \\
        \citet{wilinski2024reprsinter} & N.A. & UTS (Statistical) & Eval (Pure) & Moment/Chronos/Moirai & Representation Analysis & Yes & 2025 \\
        \citet{potosnak2025comporeason} & 100 Series & UTS (Statistical) & Eval (Mixed) & 23 models & Reasoning & Yes & 2025 \\
        Freq-Synth~\cite{nochumsohn2024freqsynth} & N.A. & MTS (Statistical) & PT (Pure)/Eval (Mixed) & 8 models & Forecasting & No & 2024 \\
        \bottomrule
    \end{tabular}%
    }
\end{table*}%

\subsubsection{Generating Synthetic Time Series in Chronos}
To enrich the pretraining corpus, Chronos~\cite{ansari2024chronos} introduces KernelSynth, a technique for generating synthetic univariate time series by randomly combining Gaussian Processes (GPs) kernels. GPs are distributions over functions, characterized by a mean function $m(t)$, and a positive definite kernel (covariance function) $\kappa(t, t^\prime)$, where $t \in \mathbb{R}$ represents the input domain. The kernel defines how values of the function at different points $t$ and $t^\prime$ are correlated. By carefully selecting kernels, diverse patterns can be generated. To this end, Chronos constructs a kernel bank $\mathcal{K}$ containing basis kernels that define fundamental time series patterns, including linear kernels for trends, periodic kernels for capturing seasonalities, and RBF kernels for smooth local variation. The final kernel $\tilde{\kappa}(t, t^\prime)$ is created by uniformly sampling kernels from $\mathcal{K}$ with replacement and combining them using random binary operations (addition or multiplication). A synthetic time series is then generated by drawing a sample of length $l_{syn}$ from the GP prior $\mathcal{GP}(m(t) = 0, \tilde{\kappa}(t, t^\prime))$.

The follow-up works either directly adopt the KernelSynth, as seen in Mamba4Cast~\cite{bhethanabhotla2024mamba4cast}, Sundial~\cite{liu2025sundial}, and Time-MoE~\cite{shi2024timemoe}, or develop new procedure based on it. For instance, TimePFN~\cite{taga2024timepfn} extends KernelSynth to generate synthetic multivariate time series by employing the linear model of coregionalization (LMC)~\cite{journel1976mininggeostat}. In LMC, outputs in each channel are derived as linear combinations of independent latent random functions produced by KernelSynth. Additionally, the number of latent functions is sampled from a Weibull distribution, while the combination weights follow a Dirichlet distribution. Notably, the LMC framework accounts for cases where correlations between different variables are weak or absent, aligning with real-world scenarios.

\subsubsection{Generating Synthetic Time Series in Moment} Moment~\cite{goswami2024moment} proposes to generate simple and fundamental univariate time series, such as sinusoidal waves, to assess whether models can learn basic patterns and to analyze models' hidden representations in response to patterns. Building on Moment's idea, \citet{potosnak2025comporeason} generate sinusoidal time series with varying frequencies and amplitudes to investigate the reasoning ability of TSFMs, while \citet{wilinski2024reprsinter} leverage similar synthetic data to identify and localize time series concepts in TSFMs. Expanding on this line of research, Freq-Synth~\cite{nochumsohn2024freqsynth} applies Fourier analysis to explore how models learn from both synthetic sine waves and real-world time series data.

\subsection{Synthetic Data Usage}
The growing fascination with synthetic time series is evident in the variety of techniques used to generate them. However, understanding how it fuels different stages of TSFM model development is just as vital. Synthetic data can act as a crucial training resource to improve model performance, a benchmarking tool for model evaluation, or a means to probe the inner workings of TSFMs. This section delves into the ways that synthetic data are harnessed, classifying existing works based on their usage of synthetic time series.

\subsubsection{Synthetic Data in Pretraining}
Most applications of synthetic data in TSFMs are concentrated in the pretraining stage. During this phase, synthetic data are either used exclusively to train the model or combined with real data. The resulting pretrained models are then applied to various downstream tasks, including forecasting and classification.

ForecastPFN~\cite{dooley2023forecastpfn}, a prior-data fitted network (PFN)~\cite{muller2022pfn}, is the first TSFM pretrained entirely on a synthetic time series corpus, enabling zero-shot downstream forecasting. Building on this, TimePFN \cite{taga2024timepfn} extends ForecastPFN to the multivariate setting, introducing the first multivariate time series PFN model pretrained on synthetic data. Mamba4Cast~\cite{bhethanabhotla2024mamba4cast} combines Mamba architecture~\cite{gu2024mamba} with PFN, achieving strong zero-shot performance while having much lower inference times than TSFMs based on the Transformer architecture. LaT-PFN~\cite{verdenius2024latpfn} innovates by integrating the PFN model with joint embedding predictive architecture frameworks, enabling in-context learning in latent space. This model is capable of both forecasting and classification. Another notable approach, TabPFN-TS~\cite{hoo2024tabpfnts}, leverages the TabPFN model, originally pretrained on synthetic tabular datasets, to perform zero-shot time series forecasting. Despite having only 11M parameters and not being trained on time series data, TabPFN-TS demonstrates impressive performance on the GIFT-Eval time series benchmark~\cite{aksu2024gift}.

While the above methods focus on designing novel model architectures, some studies take a step back and question the true value of synthetic data in pretraining. These studies yield differing conclusions. InfoBoost~\cite{fu2024infoboost} critiques existing data generation methods and presents a new approach that allows model pretraining on synthetic data to outperform models trained on real data. In contrast, \citet{kuvshinova2024synornot} reassess whether synthetic data truly enhance zero-shot forecasting quality or if relying on a limited amount of real data is sufficient. Their findings suggest that using synthetic data do not provide a performance boost in a zero-shot setting when compared to utilizing even a small set of real data.

Another line of research explores the strategy of mixing synthetic data with real data during pretraining, as seen in models like Chronos~\cite{ansari2024chronos}, TimesFM~\cite{das2024timesfm}, Sundial~\cite{liu2025sundial}, Time-MoE~\cite{shi2024timemoe}, TimeHF \cite{qi2025timehf}, and Toto~\cite{cohen2024toto}. The effectiveness of synthetic data in these approaches is thoroughly examined and validated through extensive empirical studies. For instance, Chronos systematically investigates the impact of KernelSynth on downstream model performance, varying the synthetic data proportion from 0\% to 100\%. The findings reveal that both in-domain and zero-shot performance improve with the inclusion of synthetic data, with the most consistent gains observed around the 10\% synthetic data mark. Beyond this point, increasing the proportion of synthetic data tends to degrade performance. This result aligns with the fact that data generated using GPs do not always represent the full diversity of real-world cases. Furthermore, the study confirms that models trained solely on synthetic data underperform compared to those trained with real data, supporting the conclusions drawn by \citet{kuvshinova2024synornot}.

\subsubsection{Synthetic Data in Evaluation}
Synthetic time series data are valuable tools for controllably examining the capabilities of TSFMs on basic patterns, investigating what is being learned within the models, and, in turn, helping us identify synthetic patterns that are missing from the pretraining corpus and need to be incorporated for improved performance. Take Moment~\cite{goswami2024moment} as an example. It conducts a series of experiments using synthetically generated sine waves to assess Moment's ability. Visualizations of the embeddings of these synthetic sinusoids indicate that Moment effectively captures subtle variations in trend, scale, frequency, and autocorrelation. Similarly, \citet{wilinski2024reprsinter} generate synthetic sine waves to systematically investigate how well TSFMs grasp intuitive time series concepts. Their findings reveal that certain linear concepts encoded by Moment-Large are linearly separable, though this separability is not uniform across layers but rather emerges at specific stages within the model. Beyond purely synthetic evaluations, \citet{potosnak2025comporeason} integrate sine waves with real time series data to comprehensively assess the reasoning and generalization capabilities of TSFMs. Meanwhile, Freq-Synth~\cite{nochumsohn2024freqsynth} leverages simple synthetic sine waves to design experiments that expose critical issues in existing TSFMs, particularly frequency confusion and challenges in frequency generalization.

\subsection{Limitations}
While synthetic data have played a crucial role in advancing TSFMs, current approaches still face several fundamental challenges:

\paragraph{Lack of a Systematic Approach to Incorporating Synthetic Data}
Current methods lack a structured framework for integrating synthetic data into pretraining. Ideally, a systematic approach would first analyze the pretraining corpus to identify missing patterns and then generate synthetic data specifically designed to fill those gaps. However, existing works take a more ad-hoc approach, simply creating synthetic datasets they deem important and adding them to the pretraining corpus without a clear strategy for ensuring coverage of critical missing patterns.

\paragraph{Absence of Data-Driven Generative Methods}
Most current studies rely on statistical methods, such as combining different GPs kernels, to generate synthetic time series data. While these methods offer interpretability and control, it remains unclear whether they are the most effective way to create synthetic data. Exploring data-driven generative techniques, such as diffusion models trained on real-world distributions, could potentially produce more realistic data, leading to better pretraining outcomes.

\paragraph{Missed Opportunity to Finetune with Synthetic Data}
The focus on synthetic data has almost exclusively been on pretraining, while finetuning remains an untapped potential. After a TSFM is pretrained, finetuning with carefully designed synthetic data could serve as a precise intervention to reinforce its understanding of underrepresented patterns—without the need for a costly full retraining cycle. This adaptive finetuning stage could be particularly useful for improving generalization, enhancing model robustness, and addressing domain-specific gaps.

\begin{table*}[h]
    \LARGE
    \centering
    \caption{Summary of works utilizing synthetic data for TSLLMs. N.A. indicates undisclosed information. PT: Pretraining. FT: Finetuning. Eval: Evaluation. Pure: Pure synthetic data. Mixed: Synthetic data mixed with real data. FE: Forecasting explanation.}
    \vspace{-0.5em}
    \label{tab:tsllms}%
    \resizebox{\textwidth}{!}{
    \begin{tabular}{lcccccccc}
        \toprule
        Method & \# TS-Text & TS Generation & Text Generation & Applied Stage (Purity) & Applied Model & Downstream Task & Code & Year \\
        \midrule
        \citet{chow2024tsreason} & 70K & UTS (Statistical) & Text (Template + LLM) & PT (Mixed)/FT (Mixed)/Eval (Mixed) & Mistral-7B & Captioning/Reasoning  & No & 2024 \\
        ChatTS~\cite{xie2024chatts} & 134K & MTS (Statistical) & Text (Template + LLM) & PT (Pure)/FT (Pure)/Eval (Mixed) & QWen2.5-14B-Instruct & Reasoning & Yes & 2025\\
        TempoGPT~\cite{zhang2025tempogpt} & N.A. & MTS (Statistical) & Text (Template + LLM) & PT (Pure)/FT (Pure)/Eval (Pure) & GPT2/Llama-3.2-3B & Reasoning & No & 2025 \\

        Insight Miner~\cite{zhang2023insightminer} & 10K & 
        UTS (Real) & Text (LLM) & FT (Pure)/Eval (Pure) & LLaVA & Captioning & Yes & 2023 \\
        ChatTime~\cite{wang2025chattime} & 50K & UTS (Statistical) & Text (Template) & FT (Mixed)/Eval (Mixed) & LlaMA-2-7B-Base & Forecasting/QA & Yes & 2024\\
        SysCaps~\cite{emami2024syscaps} & N.A. & UTS (Real) & Text (LLM) & FT (Pure)/Eval (Pure) & BERT/DistillBERT & Captioning & Yes & 2024 \\
        GPT4MTS~\cite{jia2024gpt4mts} & 484K & MTS (Real) & Text (Crawl + LLM) & FT (Pure)/Eval (Pure) & GPT2 & Forecasting & No & 2024 \\
        AIR~\cite{seo2024air} & N.A. & MTS (Statistical) & Text (Template) & FT (Pure)/Eval (Mixed) & TSMixer/Mistral-7B & Forecasting & No & 2024  \\
        xTP-LLM~\cite{guo2024xtpllm} & N.A. & UTS (Real) & Text (LLM) & FT (Mixed) & LLaMA2-7B-chat & Forecasting/FE & No & 2024 \\

        TRUCE~\cite{jhamtani2021truce} & 720 & UTS (Statistical) & Text (Human) & Train (Mixed)/Eval (Mixed) & Graphical Model & Captioning & Yes & 2021  \\
        TGTSF~\cite{xu2024tgtsf} & N.A. & UTS (Statistical) & Text (Template) & Train (Mixed)/Eval (Mixed) & Encoder-Decoder Transformer & Forecasting & No & 2024 \\
        \citet{fons2024evaluating} & 7.2K & MTS/UTS (Statistical) & Text (Template) & Train (Pure)/Eval (Pure) & 5 models & Reasoning & No & 2024 \\
        TSLM Caption~\cite{trabelsi2025tslm} & 188K & UTS (Data-Driven) & Text (LLM) & Train (Mixed)/Eval (Mixed) & Encoder-Decoder Transformer & Captioning & No & 2025 \\
        
        TimeSeriesExam~\cite{cai2024timeseriesexam} & 700 & UTS (Statistical) & Text (Template) & Eval (Pure) & 10 models & MCQA & Yes & 2024 \\
        XForecast~\cite{aksu2024xforecast} & 67K & UTS (Data-Driven) & Text(LLM) & Eval (Pure) & 5 models & FE & No &  2024\\
        \citet{merrill2024struggle} & 238.7K & UTS (Data-Driven) & Text (Template + LLM) & Eval (Pure) & 5 models & Forecasting/Reasoning/MCQA & Yes & 2024 \\
        LLMTime~\cite{gruver2023llmtime} & N.A. & UTS (Statistical) & Text (Template) & Eval (Mixed) & GPT-4 & Forecasting/Reasoning & Yes &  2024 \\
        Context is Key~\cite{williams2024cik} & 134 & UTS (Statistical) & Text (Human) & Eval (Mixed) & 21 models & Forecasting & Yes & 2024 \\
        Time-MMD~\cite{liu2024timemmd} & N.A. & UTS (Real) & Text (Crawl + LLM) & Eval (Mixed) & 4 models & Forecasting & Yes & 2024  \\
        TS-Reasoner~\cite{ye2024tsreasoner} & N.A.  & MTS (Data Driven) & Text (Template) & Eval (Mixed) & ChatGPT-3.5-turbo & Forecasting/Reasoning/QA & No & 2024  \\
        TESSA~\cite{lin2024tessa} & 100 & UTS (Statistical) & Text (Template) & Eval (Mixed) & GPT-4o/LLaMA3.1-8B/Qwen2-7B & Forecasting & No & 2024 \\
        \bottomrule
    \end{tabular}%
    }
\end{table*}%

\section{Synthetic Data for Time Series LLMs}
\label{sec:tsllms}
LLMs have been successfully adapted to modalities, such as vision, speech, and tabular \cite{hollmann2023tabpfn,bai2025qwen25vl,wu2024deepseekvl2,team2023gemini,lin2023videollava,huang2025stepaudio,han2023chartllama}, driven by large-scale annotated datasets that enable multimodal alignment through pretraining~\cite{chen2025sharegpt4video,kakaobrain2022coyo-700m,mei2024wavcaps}, supervised finetuning (SFT)~\cite{liu2023llava,han2023chartllama} and evaluation~\cite{li2024omnibench,wang2025charxiv,li2023m3dbench}. While TSLLMs hold similar promise~\cite{xue2023promptcast,jiang2025explainable,li2025language}, the scarcity of high-quality structured time series–text datasets impedes effective multimodal representation learning. Synthetic data address this gap by generating realistic, well-annotated pairs, facilitating multimodal alignment, improving generalization, and reducing dependence on real-world data.

Leveraging synthetic data, TSLLMs have emerged with two primary goals: enhancing classical tasks such as forecasting by integrating expert insights, leading to more accurate and domain-adapted predictions~\cite{williams2024cik,wang2025chattime,jia2024gpt4mts}, and enabling novel reasoning tasks such as time series understanding and interpretation~\cite{chow2024tsreason,ye2024tsreasoner,xie2024chatts,zhang2025tempogpt}. Similar to Section~\ref{sec:tsfm}, we next discuss synthetic data generation methods, their application, and existing limitations of TSLLMs. We provide a summary of the works discussed in this section in Table~\ref{tab:ts-gen}.

\begin{figure}[t]
    \centering
    \includegraphics[width=\linewidth]{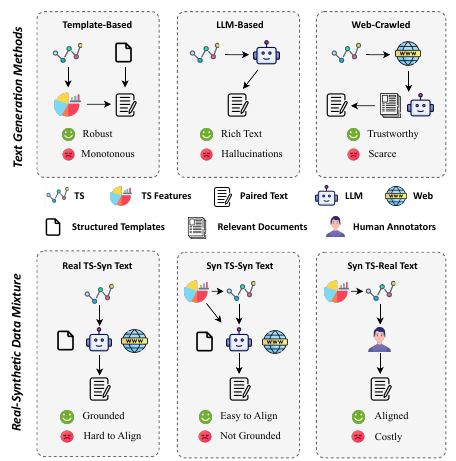}
    \caption{Two taxonomies for analyzing synthetic data in TSLLMs: the top figure categorizes based on text generation methods, while the bottom one based on the composition of synthetic and real data.}
    \label{fig:ts-generation}
\end{figure}

\subsection{Synthetic Data Generation}
The generation of synthetic data for TSLLMs varies notably in methodology and composition. While synthetic text generation methods differ substantially, approaches for generating time series are mostly uniform, relying primarily on statistical sampling from predefined distributions. Only a few works utilize data-driven methods for generating time series. Since the main challenge in TSLLMs is aligning text and time series rather than generating the time series itself, this section focuses on synthetic text generation methods and their multimodal learning implications. As shown in Figure~\ref{fig:ts-generation}, existing techniques broadly fall into three categories—template-based generation, LLM-based generation, and web-crawled data. Additionally, methods vary based on how they combine real and synthetic data: some pair real-world time series with synthetic text, while others generate both modalities synthetically.

\subsubsection{Categorization Based on Text Generation Methods}
\paragraph{Template-based approaches} These approaches generate textual descriptions by populating predefined templates with extracted time series features. Although LLMs are typically leveraged for text generation (Section~\ref{sec:preliminary_generation}), the specific challenges of grounding generation on time series often favor template-based methods, which effectively manage the limitations LLMs have with numerical data. Templates ensure consistency by explicitly linking structured text to numerical properties of time series. For instance, TimeSeriesExam~\cite{cai2024timeseriesexam} employs 104 structured templates combined with synthetic time series generation to create a benchmark for evaluating TSLLMs.  LLMTime~\cite{gruver2023llmtime} samples series from predefined distributions, pairing them with their generating distribution names to assess LLM reasoning capabilities. Similarly, ChatTime~\cite{wang2025chattime} and \citet{chow2024tsreason} generate series paired with QA pairs derived from basic characteristics such as trend and volatility. ChatTS~\cite{xie2024chatts} employs domain-specific feature taxonomies to generate series and template-based summaries. TempoGPT~\cite{zhang2025tempogpt} produces multivariate synthetic data from white-box systems, accompanied by templated reasoning questions of varying complexity. However, template-based generation lacks diversity, as it often relies on predefined structures that limit linguistic variability and domain adaptation. Since these templates do not account for dynamic interactions between time series features, they may fail to capture nuanced relationships or complex reasoning patterns.

\paragraph{LLM-based approaches} This category utilizes LLMs to generate textual descriptions conditioned on time series inputs, enabling richer and more flexible domain-specific text. For example, ChatTS \cite{xie2024chatts} introduces TSEvol, a time series instruction following a data generator. Syscaps~\cite{emami2024syscaps} combines real time series data and tabular attributes to produce paired descriptive text. Insight Miner~\cite{zhang2023insightminer} creates charts from real time series data and uses a GPT-4 Vision model to generate captions. A different approach is taken by \citet{merrill2024struggle}, who first create textual scenarios, then use GPT-4 to generate NumPy functions producing corresponding time series. TSLM Caption~\cite{trabelsi2025tslm} prompts LLMs with in-context examples to generate paired text and time series directly, followed by post-filtering to ensure quality. Lastly, XForecast \cite{aksu2024xforecast} designs a forecast explainer that integrates statistical tools with an LLM to generate explanations for real-world forecasting instances. However, these methods face challenges like hallucination and misalignment between textual and numerical data, especially critical in accuracy-dependent domains like finance and medicine where incorrect text descriptions could mislead decision-making processes.

\paragraph{Web-crawled approaches} These approaches extract textual descriptions from online sources and align them with relevant time series data, leveraging naturally occurring text from contexts like financial news, medical reports, or scientific papers. For instance, Time-MMD \cite{liu2024timemmd} crawls the web for text relevant to real time series data, then employs an LLM to filter and summarize these sources, creating paired time series–text summaries. Similarly, GPT4MTS \cite{jia2024gpt4mts} leverages an event database containing time series and corresponding textual event descriptions, crawling related articles and summarizing them using an LLM. While this yields high-quality, human-written text, challenges include the scarcity of relevant data and the difficulty of data cleaning and time series-text alignment.

\subsubsection{Categorization Based on Real-Synthetic Data Mixture}
\paragraph{Synthetic TS-Real Text}
The least common approach pairs synthetic time series with real human-annotated text, with Context is Key~\cite{williams2024cik} being the only example in our survey. This benchmark incorporates synthetic time series to improve dataset diversity while relying on human annotations for high-quality textual supervision.

\paragraph{Real TS-Synthetic Text}
One widely adopted strategy involves using real-world time series data while generating synthetic text to provide descriptions, explanations, or annotations~\cite{emami2024syscaps,zhang2023insightminer,liu2024timemmd,jia2024gpt4mts,aksu2024xforecast,chow2024tsreason}. This method benefits from the inherent realism of real-world time series, ensuring that numerical patterns reflect authentic domain-specific behaviors. However, a major challenge is maintaining a meaningful alignment between the generated text and the underlying time series, since misalignment can introduce inconsistencies that limit the model’s ability to generalize effectively.

\paragraph{Synthetic TS-Synthetic Text} To mitigate this issue, an alternative approach generates both time series and text synthetically. This method offers greater control over the data generation process, ensuring a well-defined relationship between the two modalities~\cite{gruver2023llmtime,xie2024chatts,wang2025chattime,chow2024tsreason,merrill2024struggle,trabelsi2025tslm,zhang2025tempogpt,cai2024timeseriesexam,seo2024air,aksu2024xforecast}. By explicitly designing synthetic datasets with structured dependencies, researchers can systematically evaluate model capabilities, test reasoning skills, and create benchmark datasets with known ground truth. However, the downside of fully synthetic data is the potential gap between simulated and real-world complexity, which may affect model robustness when deployed in practical applications. The choice between these strategies depends on the intended use case—leveraging real data provides authenticity but requires careful curation, while fully synthetic datasets enable controlled experimentation at the cost of real-world variability.

\begin{figure}[t]
    \centering
    \includegraphics[width=\linewidth]{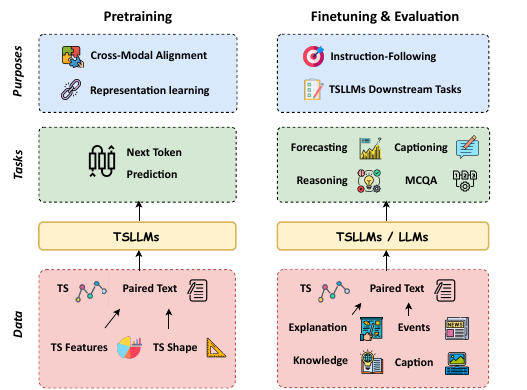}
    \caption{Illustration of the usage of synthetic data across pretraining, finetuning, and evaluation stages in TSLLMs.}
    \label{fig:ts-usage}
\end{figure}

\subsection{Synthetic Data Usage}
Beyond data generation, an equally important consideration is how the generated synthetic data are utilized in different stages of model development. From pretraining large multimodal models to finetuning and evaluating model capabilities, synthetic data play a crucial role in advancing time series–language learning. This section examines the various ways synthetic data are leveraged, categorizing existing works based on their use of synthetic data, a broad view is available in Figure~\ref{fig:ts-usage}.

\subsubsection{Synthetic Data in Pretraining} 
Pretraining acts as a bridge in multimodal learning, enabling shared representations between structurally distinct modalities like continuous numerical time series and discrete symbolic text. However, the scarcity of paired time series–language datasets makes this process challenging. Synthetic data are crucial in addressing this gap by providing large-scale training sets where time series signals are explicitly paired with descriptive text.

Several works leverage synthetic data to facilitate this alignment. ChatTS \cite{xie2024chatts} continues pretraining the QWen2.5-14B-Instruct model with an additional TS encoder using purely synthetic data, pairing univariate and multivariate synthetic time series with textual descriptions that capture their shape and local features. Whereas~\citet{chow2024tsreason} pretrain Mistral-7B again with a time series encoder and similarly structured data but sticking to univariate time series only. Moving beyond standard text-based representations, TempoGPT \cite{zhang2025tempogpt} introduces a novel approach by defining a codebook to tokenize synthetic time series into discrete tokens. These tokens are then embedded alongside textual representations, extending the LLM’s embedding space to process both modalities simultaneously during pretraining. In short, these varied approaches leverage synthetic data to create shared representation spaces, facilitating more effective multimodal learning.

\subsubsection{Synthetic Data in Finetuning}
While pretraining ensures broad alignment between time series and language, SFT enables multimodal language models to follow instructions across modalities and effectively tackle downstream tasks. Synthetic data play a vital role in this stage, allowing models to learn structured interactions between numerical data and textual prompts in a controlled setting.

All previously discussed works finetune pretrained TSLLMs using synthetic data tailored for complex reasoning and instruction-following tasks. ChatTS~\cite{xie2024chatts} generates synthetic question-answering and reasoning tasks supplemented by an instruction-following dataset based on templates. Similarly, \citet{chow2024tsreason} combine diverse synthetic downstream tasks such as captioning, QA, classification, and etiological reasoning. TempoGPT~\cite{zhang2025tempogpt} creates five synthetic reasoning tasks from its simulated environment—two simpler tasks (forecasting and trend analysis) and three complex tasks involving fault detection, diagnosis, and multivariate analysis.

Other approaches utilize synthetic data for finetuning existing TSLLMs and LLMs, or training directly on singular tasks (without applying a foundation model) resembling typical finetuning scenarios. Insight Miner \cite{zhang2023insightminer} finetunes LLaVA for time series captioning using synthetic captions generated by GPT-4. ChatTime \cite{wang2025chattime} builds on LLaMA-2-7B, finetuning it on context-guided forecasting and time series question answering, where the latter dataset is synthetically generated. TSLM Caption \cite{trabelsi2025tslm} follows a similar strategy for time series captioning but introduces a reprogramming layer on top of LLaMA2-13B, bootstrapping the training process by iteratively filtering and adding generated examples to its dataset. In contrast to these reprogramming-based approaches, GPT4MTS \cite{jia2024gpt4mts} tunes GPT-2 for the context-guided forecasting task using paired context and time series data.

\subsubsection{Synthetic Data in Evaluation}
Without meaningful benchmarks, it is difficult to assess how well a model aligns time series with language, follows instructions, or generalizes to real-world tasks. However, evaluation in this space is hindered by the scarcity of high-quality, annotated datasets that pair time series with textual descriptions, reasoning tasks, or explanations. Synthetic data once again play a crucial role in filling this gap. Evaluation is by far the most common stage where synthetic data are leveraged, serving as a key resource for assessing both TSLLMs and general-purpose LLMs on downstream tasks that require reasoning over time series and text. The nature of synthetic data used in evaluation often mirrors that of the finetuning stage, as both focus on downstream tasks rather than broad alignment, as seen in pretraining.

Most papers that employ synthetic data for finetuning also utilize it for evaluation, with some supplementing it with real data to verify performance. ChatTS \cite{xie2024chatts} is evaluated across four reasoning tasks, including inductive reasoning (in QA format), deductive reasoning (in True/False classification format), causal reasoning (in MCQA format), and comparative reasoning. Their evaluation dataset is split into two parts—one containing only real data and the other augmented with synthetic data generated using their time series–text generation pipeline. Similarly, TSLM Caption \cite{trabelsi2025tslm} evaluates its model on a mixture of real and synthetic datasets for time series captioning, using a separate synthetic data pipeline for training and evaluation. ~\citet{chow2024tsreason} assess the perception capability of its tuned LLM using the test split of its pure synthetic dataset, specifically designed for etiological reasoning, which consists of MCQA pairs that infer the most likely process generating a given time series. Insight Miner \cite{zhang2023insightminer} and ChatTime \cite{wang2025chattime} follow a similar approach, using their generation pipelines to generate pure synthetic evaluation sets for time series captioning and time series QA, respectively. Although TempoGPT~\cite{zhang2025tempogpt} also evaluates its model on the test split of its synthetic dataset, it further incorporates manual human evaluation to ensure reliable assessment of reasoning performance.

Beyond works that primarily focus on model development, several studies specialize in evaluation as their primary objective. TimeSeriesExam \cite{cai2024timeseriesexam} and \citet{merrill2024struggle} generate synthetic time series–text datasets to benchmark general-purpose LLMs on various time series understanding and reasoning tasks. Context is Key \cite{williams2024cik} adopts a hybrid approach, combining real and synthetic time series with real human-annotated text to evaluate contextual forecasting capabilities. XForecast \cite{aksu2024xforecast} generates LLM-based synthetic time series–text pairs to assess the faithfulness of explanations produced by different LLMs for time series forecasting. Finally, LLMTime \cite{gruver2023llmtime} employs synthetic time series to evaluate the etiological reasoning abilities of language models.

\subsection{Limitations}
Despite the critical role of synthetic data in enabling TSLLMs, several challenges remain.

\paragraph{Lack of Realism in Synthetic Time Series Data}
Most synthetic time series used in time series-text pairs are generated using statistical models or feature-based sampling, which often fail to capture the complexity of real-world temporal dynamics. While such approaches allow for scalable dataset generation, they may overlook important domain-specific patterns, structural dependencies, or irregular behaviors found in real-world time series. This issue is particularly significant for TSLLMs, where the interaction between text and time series is crucial—simplified synthetic time series can lead to models learning representations that do not generalize well beyond controlled experimental settings.

\paragraph{Insufficient Validation of Time Series–Text Alignment}
A key challenge in synthetic data usage is ensuring that generated time series and text are meaningfully aligned. Many current approaches use synthetic time series-text pairs across various stages—pretraining, finetuning, and evaluation—without rigorous validation of whether the text accurately describes the underlying time series patterns. This issue is particularly concerning in evaluation, where inaccurate synthetic descriptions may lead to misleading performance assessments~\cite{williams2024cik}.

\section{Future Directions}
\label{sec:future}
As synthetic data generation and usage advances, numerous promising avenues for future research and innovation emerge. This section highlights critical areas that merit deeper investigation.

\paragraph{Improving data quality and diversity} Although current approaches to synthetic data generation have demonstrated promise, there remains significant scope for enhancement. Improving the quality and diversity of synthetic data to better align with real-world data is an ongoing challenge that requires further investigation. Recent advancements of conditional time series generation ~\cite{narasimhan2024timeweaver,jing2024towards} provide more controllable way to generate customizable time series based on specific attributes. We believe that these approaches can integrate expert knowledge to ensure that the generated data align with the inherent constraints and patterns of the target domain.

\paragraph{Building human-in-the-loop synthetic data lifecycle} It will be crucial to move beyond the current one-time approach to synthetic data generation and evaluation. Incorporating a human-in-the-loop framework will be crucial for iteratively refining the synthetic data generation process. Rather than being a one-time task, synthetic data generation could become dynamic and continuous, with human feedback identifying gaps, biases, or unrealistic patterns in the generated data and correcting them in real time. This would improve the quality and relevance of the synthetic data.

\paragraph{Developing self-improvement capability} Since foundation models can themselves serve as synthetic data generation models, a natural question arises: can we leverage the synthetic data generated by a foundation model to further enhance its performance? This self-improvement capability could boost model performance by iteratively learning from the enhanced synthetic data~\cite{liu2024best}. This iterative process not only accelerates learning but also enhances the model's robustness and generalization across diverse tasks. We believe that exploring the impact of such iterative learning could reveal new avenues for foundation model development.

\section{Conclusion}
The rapid advancements in FMs have transformed the landscape of time series analysis. In this survey, we provide a thorough and up-to-date overview of how synthetic data contributes to the development of TSFMs and TSLLMs. We take a structured approach to separately examine both, delving into their methodologies through the model development lifecycle—from synthetic data generation to their usage across critical stages. Our survey offers insights into the mechanisms behind using synthetic data in both classical (e.g., forecasting) and emerging (e.g., reasoning) time series analytical tasks. By consolidating the latest advancements and identifying potential future directions, we aim to inspire innovative work in the creation of large, diverse, and high-quality datasets.


\bibliographystyle{ACM-Reference-Format}
\bibliography{sample-base}

\end{document}